\documentclass[twocolumn]{article}
\usepackage[numbers]{natbib}
\usepackage[utf8]{inputenc}
\usepackage[T1]{fontenc}
\usepackage{tgbonum}
\fontfamily{lmtt}\selectfont

\usepackage{fancyhdr}
\usepackage{authblk}
\usepackage{graphicx}
\usepackage{amsmath}
\usepackage{amssymb}
\usepackage{amsthm}
\usepackage{lipsum}
\usepackage{url}

\usepackage{algorithm}
\usepackage{algorithmic}
\usepackage{booktabs}
\usepackage{array}

\usepackage{hyperref}

\title{How Well Can Vision-Language Models Understand Humans' Intention? An Open-ended Theory of Mind Question Evaluation Benchmark}

\author {
    Ximing Wen,
    Mallika Mainali,
    Anik Sen
}

\affil{College of Computing and Informatics, Drexel University, Philadelphia, USA
\\
    
     xw384@drexel.edu,
     mm5579@drexel.edu,
     as5867@drexel.edu}

\date{}

\begin{document}

\maketitle

\thispagestyle{fancy}
\fancyhead[R]{\textit{Proc. of Int. Workshop on Advancing AI Through Theory of Mind, 2025}}
\fancyhead[L]{}
\pagestyle{fancy}
\fancyfoot{}

\begin{@twocolumnfalse}
\begin{abstract}
Vision Language Models (VLMs) have demonstrated strong reasoning capabilities in Visual Question Answering (VQA) tasks; however, their ability to perform Theory of Mind (ToM) tasks, such as inferring human intentions, beliefs, and mental states, remains underexplored. We propose an open-ended question framework to evaluate VLMs' performance across diverse categories of ToM tasks. We curated and annotated a benchmark dataset of 30 images and evaluated the performance of four VLMs of varying sizes. Our results show that the GPT-4 model outperformed all the others, with only one smaller model, GPT-4o-mini, achieving comparable performance. We observed that VLMs often struggle to infer intentions in complex scenarios such as bullying or cheating. Our findings reveal that smaller models can sometimes infer correct intentions despite relying on incorrect visual cues. The dataset is available at \url{https://github.com/ximingwen/ToM-AAAI25-Multimodal}.

\end{abstract}
\end{@twocolumnfalse}

\section*{Introduction}

Understanding human intentions through visual cues is a fundamental aspect of social intelligence, allowing effective communication, collaboration, and interaction \cite{adolphs2009social}. This capability, often referred to as the Theory of Mind (ToM), involves the ability to infer the beliefs, desires, and intentions of others based on observable behaviors and environmental contexts \cite{milligan2007language, kapogiannis2009cognitive,zhang2012perspective}. 

Recent advances in VLMs have demonstrated impressive abilities in multimodal reasoning, combining visual and textual information to perform complex tasks \cite{hessel2022androids,nagar2024zero,zhang2024large}. However, their capability to perform ToM-like reasoning, specifically in interpreting intentions from visual cues, remains underexplored. For example, \citet{etesam2023emotional} only investigate the emotional component of ToM, instead of exploring more broad categories such as intentions, religions, etc. \citet{jin2024mmtom} frame the ToM task as a binary choice question, without requiring VLMs to engage in open-ended reasoning. Consequently, this approach may not fully capture the VLMs' capability to perform ToM tasks.

To further highlight, ToM tasks present unique challenges for VLMs, requiring both visual feature extraction and contextual reasoning to infer hidden mental states. 
Thus, our study, which evaluates VLM performance on ToM tasks through an open-ended question framework, is pivotal to assessing VLMs' capacity for advanced multimodal understanding and social intelligence.

\section*{Open-ended Question Framework}

In this study, we aim to investigate the capability of VLMs to perform ToM tasks by testing their ability to interpret intentions based on visual cues in images. To fully evaluate whether VLMs truly understand humans' intentions,  we proposed an open-ended question framework composing the following three research questions:
\begin{itemize}
\item \textbf{Q1: How effectively can VLMs identify human intentions in visual scenarios?}
\item \textbf{Q2: Can VLMs recognize accurate visual cues and use them to perform ToM tasks?}
\item \textbf{Q3: Can VLMs comprehend human intentions sufficiently to make reasonable future inferences?}

\end{itemize}

\textbf{Q1} focuses on inferring individuals' mental states and intentions, a core ToM skill.
\textbf{Q2} examines the model's ability to identify and articulate visual cues, linking observations to inferred mental states.
\textbf{Q3} evaluates predictive reasoning asking the model to infer potential future actions or events based on the scene.

By designing tasks that require inferring the purpose or mental state of individuals depicted in diverse scenarios, we seek to evaluate the extent to which models align with human-like reasoning in visual intention understanding. 
Our findings contribute to research on VLMs by highlighting their strengths and limitations in approximating human cognition, paving the way for socially aware AI advancements.
\section*{Data Development}
\begin{figure*}[!h]
    \centering
    \includegraphics[width=0.95\linewidth]{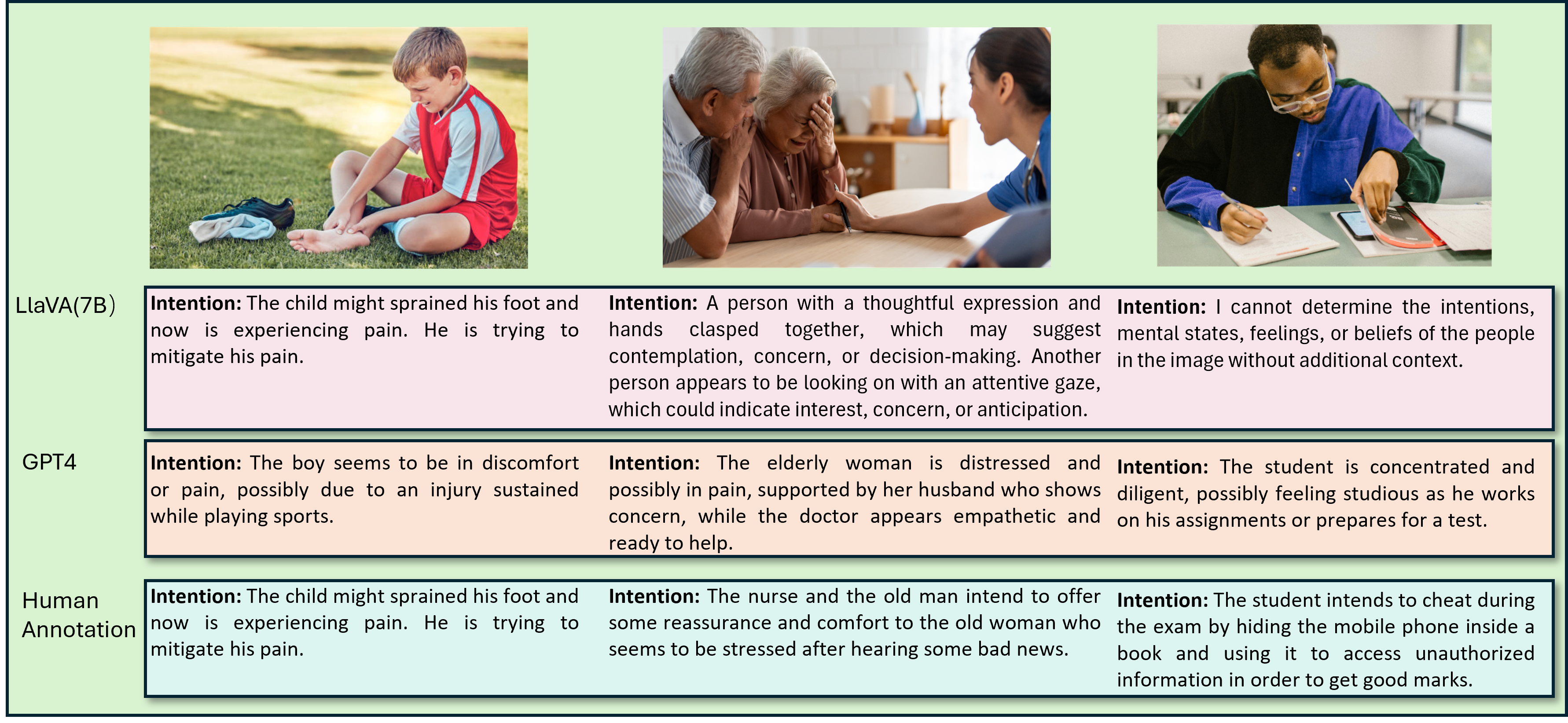} 
    \caption{An overview of the three samples used for this work to evaluate Theory of Mind (ToM) capabilities of three vision language models (VLMs). This preview shows the intention generated using two VLMs: LLaVA (7B) and GPT-4, along with the human-annotated intention.}
    \label{fig:example-image}
\end{figure*}

\noindent \textbf{Data Collection}  \quad
 We defined 30 scenarios based on two intention categories (emotion-based and action-based) and images sourced from platforms including iStock, Shutterstock, Unsplash, and Pexels under appropriate licenses to ensure copyright compliance. We only included images that conveyed clear intentions with measurable visual cues, such as facial expressions, body language, interaction with objects, and eye gaze that indicated the mental states and intentions of the individuals. Each image underwent a comprehensive review to ensure suitability for research objectives and images with ambiguous cues were excluded. The final dataset provides diverse and suitable content for research. An overview of the dataset is shown in Figure \ref{fig:example-image}.

\noindent \textbf{Data Annotation}\quad Each author annotated a subset of 10 images, providing detailed descriptions in three categories: intention, visual cues, and future inference. A third-party evaluator reviewed and validated the 30 annotations to ensure consistency and accuracy, confirming that they accurately captured intentional actions, visual cues, and potential future inferences. 

\section*{Experimental Design}
\noindent \textbf{Task} \quad
We designed a structured prompt to generate responses from VLMs that aligned with the objectives of this study. The prompt is as follows:
\textit{"Based on the given image, answer the following in one sentence each:
(1) What do you think is the intention, mental state, feeling or belief of each person in the image?
(2) What visual cues in the image helped you determine what people might be thinking or feeling?
(3) Can you infer what might happen next?"}
VLMs were expected to extract the intentions from the image, recognize visual cues that support the inferred intention, and generate plausible future scenario descriptions consistent with the context of the image.
\\

\noindent \textbf{Models}  \quad
We evaluated the performance of four VLMs - GPT-4 \cite{achiam2023gpt}, GPT-4o-mini (8B parameters) \cite{openai2024gpt4omini}, Deepseek v1 (7B parameters) \cite{bi2024deepseek}, and LLaVA (7B parameters) \cite{liu2024visual}-in inferring ToM components from images. 
\\

\noindent \textbf{Evaluation} \quad
Model responses were manually compared with human annotations using a scoring system based on keyword relevance and accuracy. A score of 1 was given for responses with correct or synonymous keywords that accurately described the context. Partially correct responses were given a score of 0.5. Smaller models, such as DeepSeek, often identified intentions correctly, but struggled with scenario details, such as misidentifying gender or objects. While object recognition errors were ignored for the `intention' category, these inaccuracies were given a score of 0.5 in the `visual cues' category. Responses without relevant keywords were given a score of 0.

\section*{Result and Discussion}
We assessed the performance of VLMs across three ToM tasks using our metric. The results, presented in Table \ref{tab:llm_comparison}, display the accuracy scores out of 30 for each category. 
\\
\noindent \textbf{Accuracy across three ToM tasks} Among the four models tested, GPT-4 performed the best on all tasks. Despite being a smaller model, GPT-4o-mini had comparable scores. In contrast, LLaVA-7B achieved significantly lower scores, indicating its limited ability to interpret subtle visual cues and make accurate inferences. Deepseek v1-7B outperformed LLaVA-7B but underperformed compared to GPT-based models. 
%

\begin{table}[h!]
\small
    \centering
    \begin{tabular}{@{}lccc@{}}
        \toprule
        \textbf{VLM} & \textbf{Intention} & \textbf{Visual Cue} & \textbf{Future Inf.} \\ 
        \midrule
        GPT4  & 27 & 27 & 28 \\
        GPT4o-mini  & 27.5 & 27  & 27.5 \\
        LLaVA-7B & 7.5 & 8.5  & 7 \\
        Deepseek-7B & 17 & 16.5  & 16  \\
      
        \bottomrule
    \end{tabular}
    \caption{Performance of VLMs' responses for inferring intention, visual cues, and future inference}
    \label{tab:llm_comparison}
\end{table}

\noindent \textbf{What types of human intentions can VLMs recognize?} \noindent
We found that GPT-based models could identify a range of human intentions, such as determination, care, frustration, compassion, praying, and bullying. Deepseek v1-7B could recognize some intentions, but struggled with subtle ones such as emotional distress, frustration, and bullying. LLaVa-7B was unable to identify most human intentions. Interestingly, none of the four VLMs could accurately identify when a person intended to cheat during an exam. They misinterpreted them as `focused' or `multitasking'. 

\noindent \textbf{Can VLMs accurately capture visual cues to infer human intentions?} 
Our analysis revealed that GPT-based models can interpret visual cues and infer human intentions, with GPT4o-mini occasionally making minor errors, such as mistaking a purse for a camera, but still capturing overall intentions accurately. However, all the four models struggled with contextual nuances, misidentifying religious attire (cassock and stole) as graduation robes, leading to incorrect inferences. 


Deepseek v1-7B could interpret body language but often misclassified facial expressions, associating direct gazes with engagement and indirect gazes with disinterest. This led to errors, such as misclassifying a police interrogation as a hospital scene, and mislabeling bullying as `amusing interaction.' It also failed to identify professions based on visible uniforms, such as firefighters and police officers, despite accurately inferring intentions. LLaVA-7B struggled to understand most visual cues.

\noindent \textbf{Can VLMs accurately interpret human intentions well enough to make reasonable future inferences?}
We found that, in most scenarios, GPT-based models made reasonable future inferences, often suggesting practical steps for conflict and emotional distress. Deepseek v1-7B also made reasonable future inferences, but its inaccuracies in intention recognition led to occasional errors. LLaVA-7B struggled with future inferences, often citing limited capabilities.

\section*{Conclusion \& Future Directions}
Our analysis shows that while some VLMs can infer human intentions from visual scenarios, they often need further fine-tuning to contextualize subtle cues. In future work, we plan to incorporate a reasoning template to guide VLMs in generating more contextually accurate responses by ensuring that key elements are considered in their reasoning process.

\section*{Acknowledgements}
The second and third authors were supported by the In the Moment (ITM) project, funded by the Defense Advanced Research Projects Agency (DARPA) under contract number HR001122S0031.

\bibliography{references}
\bibliographystyle{plainnat}

\end{document}